\documentclass{article}
\usepackage{spconf,amsmath,graphicx}
\usepackage{flushend}
\usepackage{epsfig}
\usepackage{xcolor}
\usepackage{color, colortbl}
\usepackage{cite}
\usepackage{mathrsfs}
\usepackage{subfigure}
\usepackage{subfig}
\usepackage{algorithm, algpseudocode}
\usepackage{amsmath}
\usepackage{amsthm}
\usepackage{amssymb }
\usepackage{bm}
\usepackage{multirow}  
\usepackage{booktabs}
\usepackage{enumitem}
\usepackage{pbox}
\usepackage{xfrac}


\def\ARC{\text{HDCAM}}

\def\nina{\text{Ninapro}}
\def\ConvBlock{\text{\textit{HD}Conv}}
\def\AttenBlock{\text{\textit{MHS}Atten}}
\def\ConvEnc{\text{\textit{HD}Conv Encoder}}
\def\AttenEnc{\text{\textit{MHS}Atten Encoder}}

\def\S{\text{Small}}
\def\XS{\text{XSmall}}
\def\XXS{\text{XXSmall}}

\def\W{\mathit{W}}
\def\D{\mathcal{D}}
\def\X{\bm{X}}
\def\i{_{i}}
\def\y{\mathrm{y}}

\def\L{L}
\def\C{C}

\def\R{\mathbb{R}}

\title{ Light-weighted CNN-Attention based architecture for Hand Gesture Recognition via ElectroMyography
}
\name{Soheil Zabihi$^\dagger$, Elahe Rahimian$^{\ddagger}$, Amir Asif$^\dagger$, and Arash Mohammadi$^{\ddagger}$}

\address{$^\dagger$Electrical and Computer Engineering,  Concordia University, Montreal, QC, Canada\\
$^\ddagger$Concordia Institute for Information System Engineering, Concordia University, Montreal, QC, Canada\\
}

\begin{document}
\ninept
\frenchspacing
\maketitle
%
\begin{abstract}
Advancements in Biological Signal Processing (BSP) and Machine-Learning (ML) models have paved the path for development of novel immersive Human-Machine Interfaces (HMI). In this context, there has been a surge of significant interest in Hand Gesture Recognition (HGR) utilizing Surface-Electromyogram (sEMG) signals. This is due to its unique potential for decoding wearable data to interpret human intent for immersion in Mixed Reality (MR) environments. To achieve the highest possible accuracy, complicated and heavy-weighted Deep Neural Networks (DNNs) are typically developed, which restricts their practical application in low-power and resource-constrained wearable systems. In this work, we propose a light-weighted hybrid architecture ($\ARC$) based on Convolutional Neural Network (CNN) and attention mechanism to effectively extract local and global representations of the input. The proposed $\ARC$ model with $58,441$ parameters reached a new state-of-the-art (SOTA) performance with $82.91\%$ and $81.28\%$ accuracy on window sizes of $300$ ms and $200$ ms for classifying $17$ hand gestures. The number of parameters to train the proposed $\ARC$ architecture is $18.87 \times$ less than its previous SOTA counterpart.
\end{abstract}
%
\begin{keywords}
Attention Mechanism, Biological Signal Processing (BSP), Mixed Reality (MR), surface Electromyogram.
\end{keywords}
%
\vspace{-.15in}
\section{Introduction} \label{intro}
\vspace{-.1in}
Surface Electromyogram (sEMG)-based Hand Gesture Recognition (HGR) is regarded as a promising approach for a wide range of applications, including myoelectric control prosthesis~\cite{ICASSP22_Elahe, ICASSP21_Elahe, Icassp_Elahe,Icassp}, virtual reality technologies~\cite{virtual_1,virtual_2}, Human Computer Interactions (HCI)~\cite{HCI}, and rehabilitative gaming systems~\cite{RG}. sEMG signals contain electrical activities of the muscle fibers that can be employed to decode hand gestures and thereby enhance immersive HMI wearable systems for immersion in Mixed Reality (MR) environments~\cite{3_Dario, HC-DT}. Consequently, there has been a surge of interest in the development of Deep Neural Networks (DNNs) and Machine Learning (ML) models to identify hand gestures using sEMG signals. Generally speaking, sEMG datasets can be collected based on ``sparse multichannel sEMG'' or ``High-Density sEMG (HD-sEMG)''. Despite advantages of HD-sEMG, its utilization leads to structural complexity~\cite{2021, 2022}, while adoption of sparse multichannel sEMG signals requires fewer electrodes making it the common modality of choice for  incorporation into wearable devices. Therefore, development of DNNs based on sparse sEMG signals has gained significant recent importance.

Despite extensive research in this area and the fact that academic researchers achieve high classification accuracy in laboratory conditions, there is still a gap between academic research in sEMG pattern recognition and commercialized solutions~\cite{3_Dario}. In this context, one of the objectives for reducing the gap is to focus on the development of DNN-based models that not only have high recognition accuracy but also have minimal processing complexity, allowing them to be embedded in low-power devices such as wearable controllers~\cite{ICASSP22_Elahe, Atashzar}. Furthermore, the designed DNN-based models should be based on the minimum number of electrodes while estimating the desired gestures within an acceptable delay time~\cite{3_Dario,R4}. Consequently, in this paper, we develop the novel Hierarchical Depth-wise Convolution along with the Attention Mechanism ($\ARC$) model for HGR based on sparse sEMG signals to fill this gap by meeting criteria such as improving the accuracy and reducing the number of parameters. The $\ARC$ is developed based on the $\nina$ \cite{DB1, DB1_2_3} database, which is one of the most well-known sparse multi-channel sEMG benchmark datasets.
\setlength{\textfloatsep}{0pt}
\begin{figure}[t!]
\centering
\includegraphics[width=8.6cm]{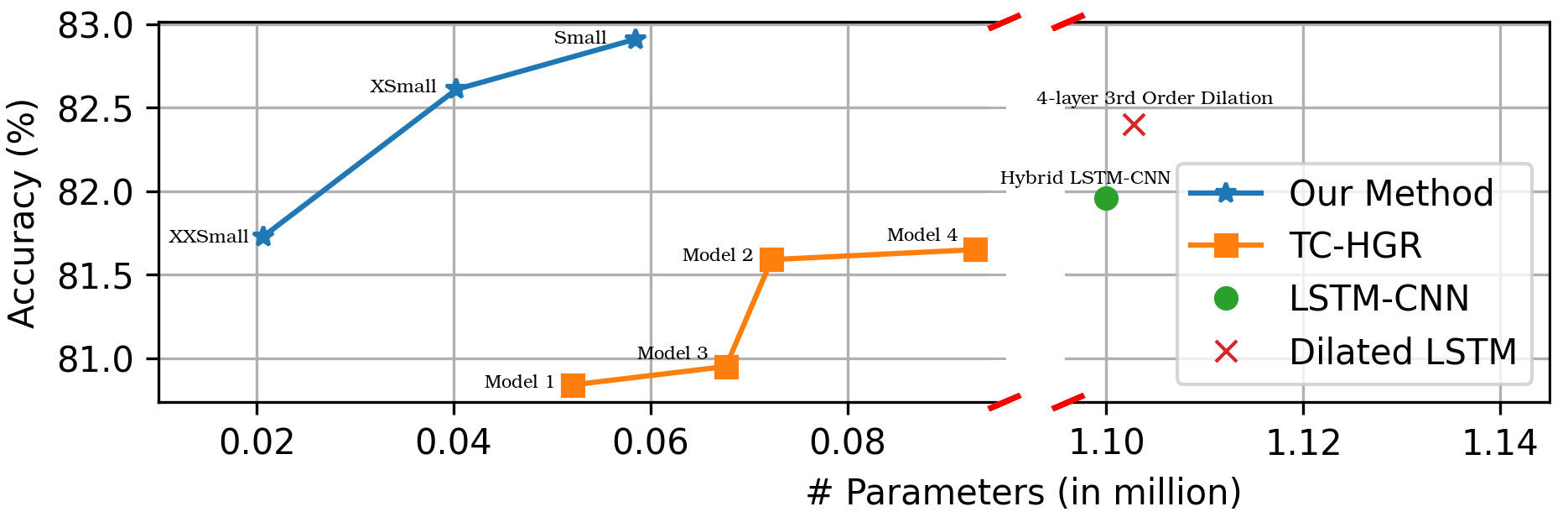}
\vspace{-0.25in}
\caption{\footnotesize Comparing different variants of our proposed $\ARC$ model with SOTA designs for an input window size of $300$ ms. The x-axis shows the number of parameters and the y-axis displays the classification accuracy on the $\nina$ DB2 dataset. Our $\ARC$ shows a better compute versus accuracy trade-off compared to recent approaches. }
\label{fig:acc_vs_param}
\end{figure}

\vspace{-.05in}
Using Convolutional Neural Networks (CNN)~\cite{GengNet, Wei2017, DingNet, WeiNet} is a common approach for hand movement classification, where sEMG signals are first converted into images and then used as input for CNN-based architectures. However, the nature of sEMG signals is sequential, and CNN architectures only  take into account the spatial features of the sEMG signals. Therefore, in recent literature~\cite{2019_letter, 2022, Atashzar}, authors proposed using recurrent-based architectures such as  Long Short Term Memory (LSTM) networks to exploit the temporal features of sEMG signals. On the other hand, it is suggested~\cite{Hybrid_2022, JMRR_Elahe, Atashzar_new} to use hybrid models (CNN-LSTM architecture) instead of using a single model to capture the temporal and spatial characteristics of sEMG signals. Although recent academic researchers are improving the performance by using Recurrent Neural Networks (RNNs) or hybrid architectures, the sequence modeling with recurrent-based architectures has several drawbacks such as consuming high memory, lack of parallelism, and lack of stable gradient during the training~\cite{Globalsip_Elahe, Icassp}. It is demonstrated~\cite{TCN} that sequence modeling using RNN-based models does not always outperform CNN-based designs. Specifically, CNN architectures have several advantages over RNNs such as lower memory requirements and faster training if designed properly~\cite{TCN}. Therefore, in the recent literature~\cite{Globalsip_Elahe, Icassp, TCN_2022, Asilomar_Elahe}, the authors took advantage of $1$-D Convolutions developed based on the dilated causal convolutions, where the sequence of sEMG signals can be processed as a whole with lower memory requirement during the training compared to RNNs. Convolution operation in CNNs, however, has two main limitations, i.e., (i) it has a local receptive field, which makes it incapable of modeling global context, and; (ii) their learned weights remain stationary at inference time, therefore, they cannot adapt to changes in input. Attention mechanism~\cite{Vaswani} can mitigate both of these problems.  Consequently, the authors in the recent research papers~\cite{Attention_2022, HuNet,ICASSP22_Elahe, TNSRE_Elahe, ICASSP21_Elahe} used the attention mechanism combined with CNNs and/or RNNs to improve the performance of sEMG-based HGR. The attention mechanism's major disadvantage is that it is often computationally intensive. Therefore, a carefully engineered design is required to make attention-based models computationally viable, especially for low-power devices.

In this paper, we develop the $\ARC$ architecture by effectively combining the complementary advantages of CNNs and the attention mechanisms. Our proposed architecture shows a favorable improvement in terms of parameter reduction and accuracy compared to the state-of-the-art (SOTA) methods for sparse multichannel sEMG-based hand gesture recognition (see Fig.~\ref{fig:acc_vs_param}). The contributions of the $\ARC$ architecture can be summarized as follows:
\begin{itemize}[leftmargin=.1in]
\vspace{-.05in}
\item  Efficiently combining advantages of Attention- and CNN-based models and reducing the number of parameters (i.e., computational burden).
\vspace{-.05in}
\item Efficiently extracting local and global representations of the sEMG sequence by coupling convolution and attention-based encoders.
\vspace{-.05in}
\item Integration of Depth-wise convolution ($DwConv$) a hierarchical structure in the proposed Hierarchical Depth-wise Convolution ($\ConvBlock$) encoder, which not only extracts a multi-scale local representation but increases the receptive field in a single block.
\vspace{-.05in}
\end{itemize}
The small version of the proposed $\ARC$ with $58,441$ parameters achieves new SOTA $82.91\%$ top-1 classification accuracy on $\nina$ DB2 dataset with $18.87$ times less number of parameters compared to the previous SOTA approach~\cite{Atashzar}.

\vspace{-.15in}
\section{The Proposed $\ARC$ Architecture} \label{architecture}
\vspace{-.1in}
\setlength{\textfloatsep}{0pt}
\begin{figure}[t!]
\centering
\includegraphics[scale=.54]{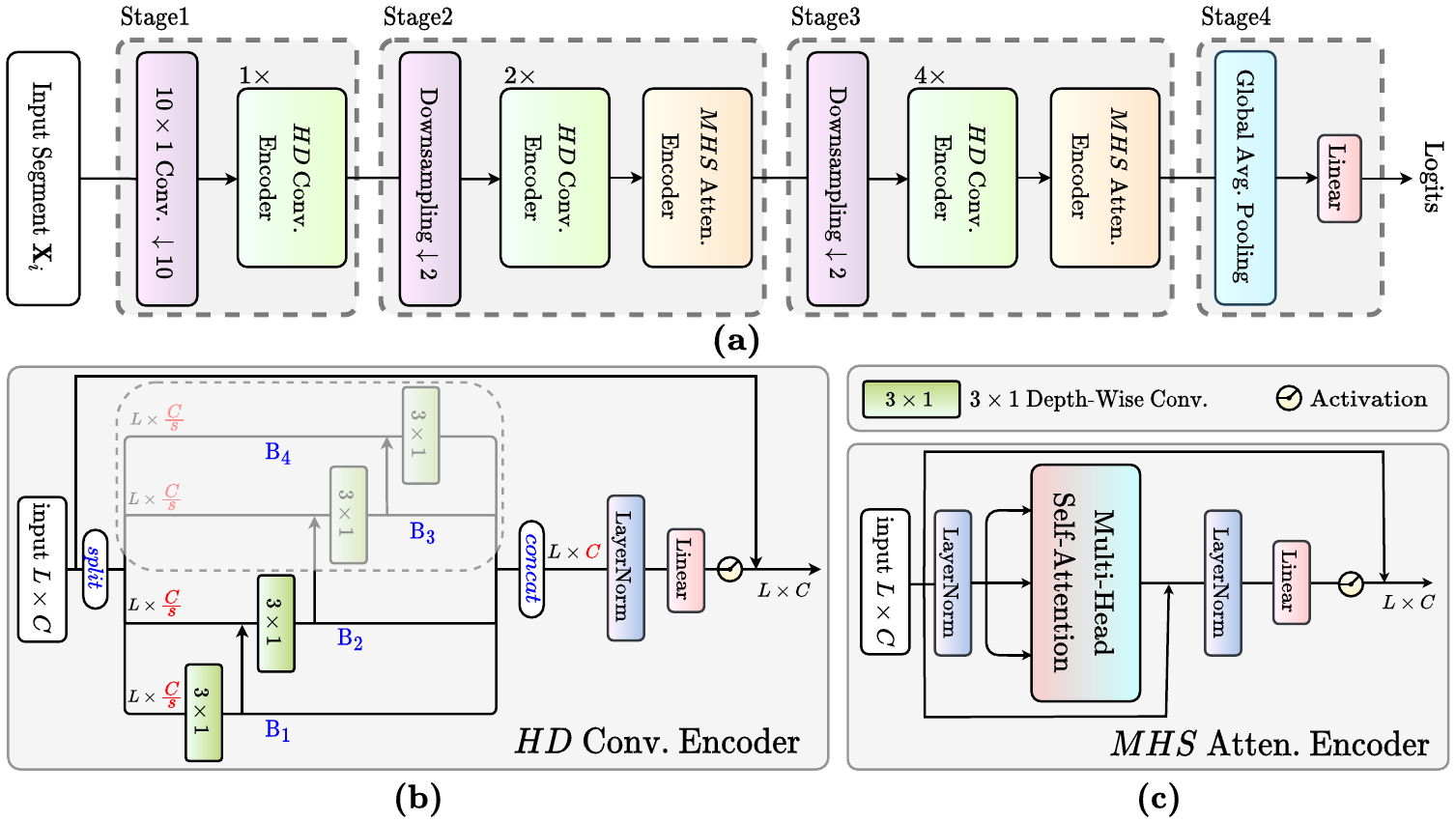}
\vspace{-0.15in}
\caption{\footnotesize \textbf{The proposed architecture:} \textbf{(a)} The overall architecture of proposed $\ARC$ model. \textbf{(b)} The $\ConvBlock$ Encoder, which uses Hierarchical Depth-wise Convolution for multi-scale temporal feature mixing. \textbf{(c)} The $\AttenBlock$ Encoder consists of a Multi-Head Self-Attention (MHA) module to encode the global representation of the input feature maps.
}
\label{fig:arc}
\end{figure}
The primary objective of this study is to build a lightweight hybrid architecture that successfully combines advantages of Attention- and CNN-based models for low-powered devices. In the proposed framework, a sliding window strategy with window of size $\W$ $\in \{150, 200, 250, 300 \, ms\}$ is adapted to use the multi-variate temporal sEMG information, resulting in dataset $\D{=}\{(\X\i, \y\i)\}_{i=1}^{N}$. More specifically, $\y_i \in \R$ is the label assigned to the $i^{\text{th}}$ segmented sequence $X_i \in \R^{L\times C}$. Here, $\L$ is the length of the segmented sequential input corresponding to the number of samples obtained at a frequency of $2$ kHz for a window of size $\W$, and $\C$ denotes the number of channels in the input segment corresponding to the number of input features/sensors. As illustrated in Fig.~\ref{fig:arc}, the $\ARC$ framework has a hybrid design based on CNN and the ``Multi-Head Self-Attention (MHA) mechanism'' to reap the advantages of both methods for designing a lightweight architecture.

\noindent
(i) \textit{\textbf{Overview of the Architecture:}} As shown in Fig~\ref{fig:arc}(a), the overall $\ARC$ architecture  consists of four different stages, the first three for multi-scale feature extraction and the last one for classification. $\ARC$ is made up of two primary components, namely ``Hierarchical Depth-wise Convolution (\ConvBlock)'' encoder and ``Multi-Head Self-Attention (\AttenBlock)'' encoder, where the former and latter aim to model the local and global information in the sequential input, respectively. Formally, for a given segmented sequential input $X_i \in \R^{L\times C}$, $\ARC$ begins by patchifing $X_i$ with a $10 \times 1$ strided convolution followed by a Layer Normalization (LN). Patching mechanism helps reduce memory and computation requirements in downstream layers resulting in ${L/10} \times C_1$ feature maps. Afterward, local features are extracted using a $\ConvBlock$ encoder. Further processing of the feature maps takes place in the second and third stages, which follow almost the same architectural structure. Both of which start with the downsampling layer followed by consecutive $\ConvBlock$ encoders for \textit{local} feature extraction and ended with $\AttenBlock$ block to encode the \textit{global} representations of the input. The downsampling layer consists of an LN followed by a $2 \times 1$ strided convolution, which reduces the sequential feature maps length by half and increases the channels, resulting ${L/20} \times C_2$ and ${L/40} \times C_3$ dimensional features for second and third stages, respectively. In the final stage, a Global Average Pooling operation is used to reduce the feature maps' dimension to $1 \times C_3$ followed by a Linear layer for classification. Here, $C_i$ refers to number of channels in $i^{\text{th}}$ stage, for $i \in \{1,2,3\}$.

\noindent
(ii) \textit{\textbf{\ConvEnc:}} As shown in Fig.~\ref{fig:arc}(b), the proposed $\ConvBlock$ block combines depth-wise convolution with a hierarchical structure to extract local features at multi-scales. The proposed multi-scale feature extractor is inspired by the Res2Net~\cite{Res2Net} module, which combines features with different resolutions. Different from the Res2Net module, we omitted the first point-wise convolution layer and added a $3 \times 1$ depth-wise convolution to the first branch. Also, the number of active branches in the hierarchical convolutional structure is dynamic and varies depending on the stage. In $\ConvBlock$ module, input feature maps of shape $L\times C$ is evenly splitted into $s$ subsets/scales, denoted by $\bm{x}\i$ of shape $L\times C/s$, where $i \in \{1,2,\dots,s\}$. Then, $3 \times 1$ depth-wise convolutions, denoted by $DwConv_i$, is applied on each subset $\bm{x}\i$ after combining with the previous branch output features, denoted by $\bm{y}_{i-1}$. Generally, we can write the output features of each branch $\bm{y}_i$ as follows
\vspace{-0.1in}
\begin{equation}
  \bm{y}_i =
    \begin{cases}
      DwConv_i(\bm{x}_i) & i=1\\
      DwConv_i(\bm{x}_i + \bm{y}_{i-1}) & 2 \leq i \leq s\\
    \end{cases}
    \label{HDWConv}
    \vspace{-0.1in}
\end{equation}
As shown in Fig~\ref{fig:arc}(b) and Eq.~\eqref{HDWConv}, the hierarchical structure allows each depth-wise convolution $DwConv_i$ receive the information from all previous splits, \{$\bm{x}_j, j \leq i$\}. The output feature maps of all branches are concatenated and passed through an LN followed by point-wise convolution to enrich the multi-scale local representation, and finally Gaussian Error Linear Unit (GELU) activation is used for adding non-linearity to the model. For information flow through the network hierarchy, residual connection is used in $\ConvBlock$ encoder. The $\ConvBlock$ encoder can be represented as follows
\vspace{-.1in}
\begin{equation}
  X_{out} = X_{in} + Linear_{GELU}(LN(HDwConv(X_{in}))),
\vspace{-.1in}
\end{equation}
where $X_{in}$ and $X_{out}$ are the $\ConvBlock$ input and output feature maps, both of shape $L \times C$, $Linear_{GELU}$ is point-wise convolution followed by GELU non-linearity $LN$ is Layer Normalization, and $HDwConv$ is hierarchical depth-wise convolution operation.

\noindent
(iii) \textit{\textbf{$\AttenBlock$ Encoder:}} In~\cite{Vaswani}, the authors showed that the attention mechanism allows a model to present global information in a given input sequence. Furthermore, attention-based architectures~\cite{Attention_2022, HuNet,ICASSP22_Elahe, TNSRE_Elahe, ICASSP21_Elahe} have shown promising performance in the context of sEMG-based HGR by extracting particular bits of information from the sequential nature of the sEMG signals. However, most of these models are still heavy-weight to be used in resource-constrained devices. Hence, in the proposed $\ARC$ architecture, we designed a hybrid architecture that combines convolutions and attention mechanism advantages. Specifically, due to spatial inductive biases in convolution operation, the CNN-based encoder ($\ConvBlock$) assists our hybrid model to learn local representations with fewer parameters than solely attention-based models. However, to effectively
learn global representations, we also used an attention-based encoder ($\AttenBlock$). Since computation in the MHA has quadratic relation to input size, we only used the $\AttenBlock$ encoder in the second and third stages of the $\ARC$ to efficiently encode the global representation, where the length of the sequential feature maps are $1/20$ and $1/40$ of the original input of the network, respectively. The $\AttenBlock$ encoder can be represented as follows
\vspace{-.075in}
\begin{equation}
  X_{out} = Linear_{GELU}(LN(X_{in} + M\!H\!A(LN(X_{in})))) + X_{in}
\vspace{-.075in}
\end{equation}
where $X_{in}$ and $X_{out}$ are the $\AttenBlock$ input and output feature maps, both of shape $L \times C$, $Linear_{GELU}$ is point-wise convolution followed by GELU non-linearity $LN$ is Layer Normalization, and $M\!H\!A$ is Multi-Head Self-Attention mechanism.
In $M\!H\!A$, the input feature maps $X_{in}$ of shape $L\times C$ are passed through a Linear projection to create Queries $Q$ , i.e., a matrix with the same shape as the input feature maps. Then, Queries $Q$ is evenly splitted into $h$ subsets, denoted by $\bm{q}_i$ of shape $L\times C/h$, where $i \in \{1,2,\dots,h\}$ and $h$ is number of the heads. In parallel, the same approach has been applied to construct Keys and Values subsets, i.e, $\bm{k}_i$ and $\bm{v}_i$. Finally, on each head, the attention block measures the pairwise similarity of each $\bm{q}_i$ and all $\bm{k}_h$ to assign a weight to each $\bm{v}_h$. The entire operation is
\vspace{-0.1in}
\begin{equation}\label{attention}
    A_h = Softmax(\frac{\bm{q}_h\bm{k}_h^T}{\sqrt{d}})\bm{v}_h,
\vspace{-0.15in}
\end{equation}
where $\mathit{d}{=}C/h$ denotes the dimension of $\bm{k}_h$ and $\bm{q}_h$ subsets. Then concatenation of attention feature maps of all heads is projected to get the final attention maps of the MHA mechanism, i.e., $M\!H\!A(X_{in}){=}Linear(Concat(A_1, A_2,\dots,A_h))$.

\vspace{-.15in}
\section{Experiments and Results} \label{results}
\vspace{-.1in}

\vspace{.025in}
\noindent
\textbf{3.1. Database} \label{database}

\noindent
The proposed $\ARC$ is trained and tested using the second $\nina$ dataset~\cite{DB1_2_3} referred to as the DB2, which is the most commonly used sparse sEMG benchmark. In the DB2 dataset, muscle electrical activities are measured by the Delsys Trigno Wireless EMG system with $12$  electrodes at a  frequency rate of $2$ kHz. The DB2 dataset consists of $50$ different hand movements (including rest) recorded from $40$ healthy users. Each gesture is repeated six times by each user, where each repetition lasts for $5$ seconds,  followed by $3$ seconds of rest. For a fair comparison, as well as following the recommendations of the database~\cite{DB1_2_3} and previous literature~\cite{TNSRE_Elahe, ICASSP22_Elahe, Atashzar, Atashzar_new}, we considered two repetitions (i.e., $2$ and $5$) for testing and the rest for training. The DB2 dataset is presented in three sets of exercises (B, C, and D). Following~\cite{ICASSP22_Elahe, Atashzar, Atashzar_new}, the focus is on Exercise B which consists of $17$ hand movements. 

\begin{table}[t!]
\small
\centering
\renewcommand\arraystretch{1.72}
\caption{\footnotesize $\ARC$ Architecture  variants. Description of the models' layers with respect to kernel size, and output channels, repeated $n$ times. We use a hierarchical structure in $\ConvEnc$ to extract multi-scale local features. Also, $\AttenEnc$ is used to extract global representations of the feature maps.\label{tab:params}}
\vspace{-.25in}
\setlength{\tabcolsep}{5pt}
\begin{center}
\scalebox{0.68}{
\begin{tabular}{l l c c l l l}
\hline
\hline
& \multirow{2}{*}{Layer} &  \multirow{2}{*}{\#Layers ($n$)} & \multirow{2}{*}{Kernel Size} & \multicolumn{3}{c}{Output Channels} \\
\cline{5-7}
 &   &   &  & \textbf{\XXS} & \textbf{\XS} & \textbf{\S} \\
\hline
\multicolumn{1}{c}{\multirow{2}[1]{*}{\rotatebox[origin=c]{90}{ Stage 1}}}
& Stem              & $1$  &  $10\times1$  & $16$ & $24$ & $24$\\
& \ConvEnc    & $1$  &   $3\times1$  & $16(s=2)$ & $24(s=3)$ & $24(s=3)$\\
\hline
\multicolumn{1}{c}{\multirow{3}[1]{*}{\rotatebox[origin=c]{90}{Stage 2}}}
& Downsampling      & $1$  &   $2\times1$  & $24$ & $32$ & $32$\\
& \ConvEnc    & $2$  &   $3\times1$  & $24(s=3)$ & $32(s=4)$ & $32(s=4)$\\
& \AttenEnc    & $1$  &   $-$           & $24(h=3)$ & $32(h=4)$ & $32(h=4)$\\
\hline
\multicolumn{1}{c}{\multirow{3}[1]{*}{\rotatebox[origin=c]{90}{Stage 3}}}
& Downsampling      & $1$  &   $2\times1$  & $32$ & $48$ & $64$\\
& \ConvEnc    & $4$ &   $3\times1$  & $32(s=4)$ & $48(s=4)$ & $64(s=4)$\\
& \AttenEnc    & $1$  &   $-$           & $32(h=4)$ & $48(h=4)$ & $64(h=4)$\\
\hline
& Global Avg. Pooling   & $1$                           &  $-$ & $32$ & $48$ & $64$\\
& Linear                & $1$                           &  $1$ & $17$ & $17$ & $17$\\
\hline
&\textbf{Model Parameters}   &&  & $20,689$ & $40,281$ & $58,441$\\
\bottomrule
\hline
\hline
\end{tabular}}
\vspace{-.15in}
\end{center}
\end{table}

\vspace{.025in}
\noindent
\textbf{3.2. Results and Discussions} \label{Results}

\noindent
In this section, a comprehensive set of experiments is conducted to evaluate the performance of the proposed $\ARC$ architecture. Table~\ref{tab:params} represents the sequence of the $\ConvBlock$ and $\AttenBlock$ encoders along with design information of the extra-extra small (\XXS), extra-small (\XS), and small (\S) versions of the model. As shown in Table~\ref{tab:params}, the type, number, and sequence of the component blocks in the overall model architecture (illustrated in Fig.~\ref{fig:arc}) are maintained across all $\ARC$ architecture variants. The number of output channels in each stage is what distinguishes the $\XXS$, $\XS$, and $\S$ models from one another. Since the number of active branches ($s$) and attention heads ($h$) in the $\ConvBlock$ and $\AttenBlock$ encoders is proportional to the number of output channels of the corresponding stage, we maintained the fundamental rule for all model variants, which requires having at least eight channels per-head/per-branch. Additionally, the maximum allowed number of heads/branches is set to four. All models were trained using the Adam optimizer at a learning rate of $0.0001$ with Cross-Entropy loss. Furthermore, we chose a mini-batch size of $32$.
For the pre-processing step, we employed the $1^{\text{st}}$ order $1$ Hz low-pass Butterworth filter to smooth raw sEMG signals as described in the recent literature~\cite{pattern_letter2019,GengNet, AtzoriNet,DB1_2_3}. Moreover, we normalized and scaled the sEMG signals by the \textit{$\mu$-law} technique~\cite{Icassp_Elahe} with the \textit{$\mu$} value of $256$.
Table~\ref{tab:internalCompare} shows the average recognition accuracy of different variants over all subjects. In the following sections, we focus on five experiments to evaluate our proposed
$\ARC$ model:

\noindent
\textbf{\textit{The Model's Dimension:}} This experiment analyzes the recognition accuracy of the $\ARC$  by varying the number of channels in each stage, yielding $\XXS$, $\XS$, and $\S$ models. In this regard, Table~\ref{tab:internalCompare} shows the results for all variants of the proposed architecture for different window sizes. For the same arrangement of component layers, it can be seen from Tables~\ref{tab:params} and~\ref{tab:internalCompare} that the accuracy of the model is improved by increasing the dimensions of the stages.
More specifically, the dimension of the stage $3$ is the only difference between the $\XS$ and $\S$ architectures, resulting in more informative high-level features in the $\S$ model, which leads to better performance. Comparing $\XXS$ versus two other variants, the dimension of all stages has reduced leading to lower performance (Table~\ref{tab:internalCompare}).  From Table~\ref{tab:params}, it can be observed that there is a trade-off between the complexity of the model and the accuracy.

\noindent
\textbf{\textit{The Effect of Window Size:}} Comparing outcomes in each column of  Table~\ref{tab:internalCompare} shows that increasing window size ($W$) led to better performance for all model variants. It is worth mentioning that $W$ is required to be under $300$ ms to have a real-time response in peripheral human machine intelligence systems~\cite{R4}. According to this observation, the proposed $\ARC$ architecture is capable of extracting/utilizing information from longer sequences of inputs.
Larger $W$ leads to longer sequential feature maps in the second and third stages, which leads to more memory requirements in the attention mechanism. We would like to emphasize that our proposed hybrid architecture is still far superior to the sole attention-based approach since the sequence lengths in the second and third stages are significantly decreased, as previously noted.

\begin{table}[t!]
\small
\centering
\renewcommand\arraystretch{1.3}
\caption{\footnotesize Accuracy of $\ARC$ variants over different window sizes ($\W$).\label{tab:internalCompare}}
\vspace{-.1in}
\scalebox{0.72}{
{\begin{tabular}{c c c c c}
\hline
\hline

& \multicolumn{1}{c}{\textbf{Model ID}}
& \textbf{\XXS}
& \textbf{\XS}
& \textbf{\S}
\\
\cline{2-5}
\multicolumn{1}{c}{\multirow{2}[1]{*}{\rotatebox[origin=c]{90}{\footnotesize $\W$=}}} \newline \multirow{2}[1]{*}{\rotatebox[origin=c]{90}{\footnotesize 150 ms}}
&
\multicolumn{1}{c}{Accuracy ($\%$)} & $80.53$   & $81.51$   & $82.21$
\\
&
\multicolumn{1}{c}{STD ($\%$)}      & $6.9$     & $6.6$     & $6.7$
\\
\hline
\multicolumn{1}{c}{\multirow{2}[1]{*}{\rotatebox[origin=c]{90}{\footnotesize $\W$=}}} \newline \multirow{2}[1]{*}{\rotatebox[origin=c]{90}{\footnotesize 200 ms}}
&
\multicolumn{1}{c}{Accuracy ($\%$)} & $81.10$   & $81.77$   & $82.28$
\\
&
\multicolumn{1}{c}{STD ($\%$)}      & $6.8$     & $6.8$     & $6.6$
\\
\hline
\multicolumn{1}{c}{\multirow{2}[1]{*}{\rotatebox[origin=c]{90}{\footnotesize $\W$=}}} \newline \multirow{2}[1]{*}{\rotatebox[origin=c]{90}{\footnotesize 250 ms}}
&
\multicolumn{1}{c}{Accuracy ($\%$)} & $81.26$   & $82.17$   & $82.57$
\\
&
\multicolumn{1}{c}{STD ($\%$)}      & $6.8$     & $6.7$     & $6.6$
\\
\hline

\multicolumn{1}{c}{\multirow{2}[1]{*}{\rotatebox[origin=c]{90}{\footnotesize $\W$=}}} \newline \multirow{2}[1]{*}{\rotatebox[origin=c]{90}{\footnotesize 300 ms}}
&
\multicolumn{1}{c}{Accuracy ($\%$)} & $81.73$   & $82.61$   & $82.91$
\\
&
\multicolumn{1}{c}{STD ($\%$)}      & $6.7$     & $6.6$     & $6.5$
\\
\hline
\hline
\end{tabular}}}
\vspace{0in}
\end{table}

\noindent
\textbf{\textit{Comparison with State-of-the-Art (SOTA):}}
In Table~\ref{tab:externalCompare}, $\ARC$ is compared with recent SOTA recurrent (Dilated LSTM)~\cite{Atashzar}, convolutional (CNN), hybrid LSTM-CNN~\cite{Atashzar_new}, and hybrid attention-CNN~\cite{ICASSP22_Elahe} models on Ninapro DB2 dataset~\cite{DB1_2_3}. Overall, our model demonstrates better accuracy versus the number of parameters compared to other methods. As shown in Table~\ref{tab:externalCompare}, for the window size of $200$ ms, all variants of the proposed model outperform other SOTA approaches. For instance, our $\XXS$ model has  $53.3$ times less parameter than Dilated LSTM, but obtains a $2.1\%$ gain in the top-$1$ accuracy. Compared to the best performing TC-HGR model (Model $4$), our $\XXS$ and $\S$ models improve the accuracy for $0.38\%$ and $1.56\%$ with $4.59$ and $1.62$ times less number of parameters, respectively. Moreover, as shown in Table~\ref{tab:externalCompare}, for the window size of $300$ ms, $\XS$ and $\S$ variants of $\ARC$ obtain $82.61\%$ and $82.91\%$ top-1 accuracy respectively, both surpassing all previous SOTA methods with fewer number of parameters (see Fig.~\ref{fig:acc_vs_param}). Dilation-based LSTM~\cite{Atashzar}, as the previous SOTA model on DB2 dataset, reached to $82.4\%$ top-1 accuracy with $1,102,801$ number of parameters, while our $\XS$ model attains better accuracy ($82.61\%$) with only $40,281$ parameters, i.e., $27.38$ times fewer. It is worth noting that our $\S$ model achieved $82.91\%$ top-1 accuracy with $58,441$ parameters, reaching a new SOTA performance that demonstrates the effectiveness and the generalization of our design. 

\noindent
\textbf{\textit{Effectiveness of the Multi-scales Local Representation:}} To extract multi-scales local features, we integrated depth-wise convolution ($DwConv$) with a hierarchical structure in the proposed $\ConvBlock$ encoder. The hierarchical structure besides the multi-scale feature extraction increases the receptive field in a single block. As shown in Table~\ref{tab:no_hira}, replacing the \textit{``hierarchical''} $DwConv$ structure in $\ConvBlock$ with a standard $DwConv$ layer degrades the accuracy in all variants of $\ARC$, indicating its usefulness in our design. As an example, the top-1 accuracy of the $\S$ model decreased by $0.56\%$ in its non-hierarchical variant.

\begin{table}[t!]
\centering
\renewcommand\arraystretch{2}
\caption{\footnotesize Performance comparison with SOTA models. \label{tab:externalCompare}}
\vspace{-.15in}
\scalebox{.58}{
{\begin{tabular}{l c c c c c}
\hline
\hline
& \multirow{2}[2]{*}{\textbf{Models}}
& \multicolumn{2}{c}{\textbf{$\bm{W = 200}$ ms}}
& \multicolumn{2}{c}{\textbf{$\bm{W = 300}$ ms}}
\\
\cline{3-6}
&
& \multicolumn{1}{c}{\textbf{Parameters$\downarrow$}}
& \textbf{Accuracy$\uparrow$ ($\%$)}
& \multicolumn{1}{c}{\textbf{Parameters$\downarrow$}}
& \textbf{Accuracy$\uparrow$ ($\%$)}
\\
\hline
\multicolumn{1}{c}{\multirow{2}[2]{*}{\rotatebox[origin=c]{90}{\footnotesize \textbf{\:\;\;Dilated}}}} \newline \multirow{2}[2]{*}{\rotatebox[origin=c]{90}{\footnotesize \textbf{$\:\;\;$LSTM}~\cite{Atashzar}}}
& \multicolumn{1}{c}{\footnotesize 4-layer 3rd Order Dilation}
& $1,102,801$
& $79.0$
& $1,102,801$
& $82.4$
\\
& {\footnotesize 4-layer 3rd Order Dilation (pure LSTM)}
& $\_$
& $\_$
& $466,944$
& $79.7$
\\
\hline

\multicolumn{1}{c}{\multirow{2}[2]{*}{\rotatebox[origin=c]{90}{\footnotesize \textbf{\:\;\;LSTM$-$}}}} \newline \multirow{2}[2]{*}{\rotatebox[origin=c]{90}{\footnotesize \textbf{\:\;\;CNN}~\cite{Atashzar_new}}}
& \multicolumn{1}{c}{CNN}
& -
& -
& $\approx 1.4M$
& $77.30$
\\
& \multicolumn{1}{c}{Hybrid LSTM-CNN}
& -
& -
& $\approx 1.1M$
& $81.96$
\\
\hline

\multicolumn{1}{c}{\multirow{4}[2]{*}{\rotatebox[origin=c]{90}{\footnotesize \textbf{\:\;\;TC-HGR}}}} \newline \multirow{4}[2]{*}{\rotatebox[origin=c]{90}{\footnotesize \textbf{\:\;\;\cite{ICASSP22_Elahe}}}}
& \multicolumn{1}{c}{Model 1}
& $49,186$
& $80.29$
& 52,066
& $80.84$
\\
& \multicolumn{1}{c}{Model 2}
& $68,445$
& $80.63$
& $72,285$
& $81.59$
\\
& \multicolumn{1}{c}{Model 3}
& $69,076$
& $80.51$
& $67,651$
& $80.95 $
\\
& \multicolumn{1}{c}{Model 4}
& $94,965$
& $80.72$
& $92,945$
& $81.65$
\\
\hline

\multicolumn{1}{c}{\multirow{3}[2]{*}{\rotatebox[origin=c]{90}{\footnotesize \textbf{\:\;\;Ours}}}} \newline \multirow{3}[2]{*}{\rotatebox[origin=c]{90}{\footnotesize \textbf{\:\;\;(\;$\ARC$\;)}}}
& \multicolumn{1}{c}{\XXS}
& $\textbf{20,686}$
& $81.10$
& \textbf{20,686}
& $81.73$
\\
& \multicolumn{1}{c}{\XS}
& $40,281$
& $81.77$
& $40,281$
& $82.61$
\\
& \multicolumn{1}{c}{\S}
& $58,441$
& $\textbf{82.28}$
& $58,441$
& $\textbf{82.91}$

\\
\hline
\hline
\end{tabular}}}
\vspace{-.1in}
\end{table}
\begin{table}[t!]
\small
\centering
\renewcommand\arraystretch{1.2}
\caption{\footnotesize Evaluating the effectiveness of  the multi-scales local representation extraction in the $\ConvBlock$ encoder. For Hierarchical models, Table~\ref{tab:params} shows the scale values ($s$) for each stage, otherwise, $s$ is equal to $1$ at all stages.  \label{tab:no_hira}}
\vspace{-.1in}
\scalebox{.74}{
{\begin{tabular}{  l  c c c}
\hline
\hline
\multirow{2}[2]{*}{\textbf{$\ConvEnc$}}
            & \multicolumn{3}{c}{\textbf{Accuracy ($\%$)}}
\\
\cline{2-4}
& \XXS
& \XS
& \S
\\
\hline

Hierarchical structure
& $81.73$
& $82.61$
& $82.91$
\\
Non-hierarchical structure
& $81.27$
& $82.21$
& $82.35$
\\
\hline
\hline
\end{tabular}}}
\vspace{-0.1in}
\end{table}

\noindent
\textbf{\textit{Importance of Using $\AttenBlock$ Encoders:}} To examine the importance of $\AttenBlock$ encoder, we conducted two ablation studies using this encoder at different stages of the network for $W{=}300$ ms. In Table~\ref{tab:AttenBlock}, we kept the total number of $\ConvBlock$ and $\AttenBlock$ encoders fixed to $[1,3,5]$ for experiment $1$ to $4$. While in experiment $5$ to $8$, the number of $\ConvBlock$ encoder is set to $[1,2,4]$ for all stages, and $\AttenBlock$ encoder is progressively added to the end of stages.  According to both experiments, adding the $\AttenBlock$ encoder gradually in the last two stages increases accuracy($\checkmark$) and the number of parameters($\bm{\times}$). In addition, adding a global $\AttenBlock$ encoder to the first stage is not beneficial since the features in this stage are not mature enough.
%
Furthermore, we conducted another experiment to investigate the impact of using the $\AttenBlock$ encoder at the beginning (after downsampling) versus end of each stage on the $\ARC$ architecture. As shown in Table~\ref{tab:AttenBlockPos}, better performance is achieved by using the $\AttenBlock$ encoder as the final block of the stages.

\begin{table}[t!]
\small
\centering
\renewcommand\arraystretch{1.3}
\caption{\footnotesize Evaluating impact of $\AttenBlock$ encoder at different stages of the network. The listed values show the number of the corresponding encoder in stages $1$ to $3$ in order. Highlighted rows indicate the $\S$ model.  \label{tab:AttenBlock}}
\vspace{-.12in}
\scalebox{0.74}{
{\begin{tabular}{r l c c}
\hline
\hline
\textbf{ID:}
& \multicolumn{1}{l}{\textbf{Model Configuration}}
            & \textbf{Accuracy$\uparrow$ ($\%$)}
            & \textbf{Parameters$\downarrow$}
\\
\hline
$\bm{1:}$
&$\ConvBlock=[1,3,5], \AttenBlock=[0,0,0]$
& $81.56$
& $37,673$
\\
$\bm{2:}$
&$\ConvBlock=[1,3,4],\AttenBlock=[0,0,1]$
& $82.45$
& $54,249$
\\
\rowcolor{blue!6}
$\bm{3:}$
&$\ConvBlock=[1,2,4],\AttenBlock=[0,1,1]$
& $82.91$
& $58,441$
\\
$\bm{4:}$
&$\ConvBlock=[0,2,4],\AttenBlock=[1,1,1]$
& $82,26$
& $60,817$
\\
\hline
$\bm{5:}$
&$\ConvBlock=[1,2,4], \AttenBlock=[0,0,0]$
& $81.93$
& $31,785$
\\
$\bm{6:}$
&$\ConvBlock=[1,2,4],\AttenBlock=[0,0,1]$
& $82.55$
& $52,969$
\\
\rowcolor{blue!6}
$\bm{7:}$
&$\ConvBlock=[1,2,4],\AttenBlock=[0,1,1]$
& $82.91$
& $58,441$
\\
$\bm{8:}$
&$\ConvBlock=[1,2,4],\AttenBlock=[1,1,1]$
& $82.48$
& $61,585$
\\
\hline
\hline
\end{tabular}}}
\vspace{-.1in}
\end{table}
\begin{table}[t!]
\small
\centering
\renewcommand\arraystretch{1.1}
\caption{\footnotesize Evaluating the impact of using $\AttenBlock$ encoder at the beginning vs. end of each stage for the window size of $300$ ms. \label{tab:AttenBlockPos}}
\vspace{-.13in}
\scalebox{0.74}{
{\begin{tabular}{  l  c c c}
\hline
\hline
\multirow{2}[2]{*}{\textbf{$\AttenEnc$}}
            & \multicolumn{3}{c}{\textbf{Accuracy ($\%$)}}
\\
\cline{2-4}
& \XXS
& \XS
& \S
\\
\hline

Fist block of Stage ($\AttenBlock=[0, 1, 1]$)
& $81.31$
& $82.37$
& $82.70$
\\
Latest block of Stage ($\AttenBlock=[0, 1, 1]$)
& $81.73$
& $82.61$
& $82.91$
\\
\hline
\hline
\end{tabular}}}
\vspace{0.05in}
\end{table}
\vspace{-.15in}
\section{Conclusion}\label{sec:page}
\vspace{-.1in}

In this paper, a novel resource-efficient architecture, referred to as the $\ARC$, is developed for HGR from sparse multichannel sEMG signals. $\ARC$ is developed by effectively combining the advantages of Attention-based and CNN-based models for low-powered devices, which is an important step toward incorporating DNN models into wearable for immersive HMI. To efficiently extract local and global representations of the input sEMG sequence, $\ARC$ is empowered with convolution and attention-based encoders, namely $\ConvBlock$ and $\AttenBlock$, respectively. Specifically, we showed that by proper design of convolution-based architectures, we not only can extract a multi-scale local representation but also can increase the receptive field in a single block. A path for future study is to address the issue of misplacement or relocation of sEMG electrodes/sensors, which is an active field of research. This may be seen as both a constraint for the current study and a promising area for future investigation, which is the subject of our continuing research.


\end{document}